\documentclass[letterpaper, 10 pt, conference]{ieeeconf}  
\let\labelindent\relax
\pdfoutput=1 

\usepackage{soul}
\usepackage{epsfig}
\usepackage{graphicx}
\usepackage{amsmath}
\usepackage{amssymb}
\usepackage{ntheorem}
\usepackage{color}
\usepackage{array,multirow}
\usepackage{algorithm}
\usepackage{accents}
\usepackage{enumitem}
\usepackage{algorithmicx}
\usepackage[noend]{algpseudocode}
\usepackage{subfigure}
\usepackage[export]{adjustbox}
\usepackage{cite}

\usepackage{booktabs}

\theoremstyle{plain}
\newtheorem{defn}{Definition}

\algnewcommand\algorithmicinput{\textbf{Input:}}
\algnewcommand\INPUT{\item[\algorithmicinput]}
\algnewcommand\algorithmicpose{\textbf{Pose disambiguation:}}
\algnewcommand\Pose{\item[\algorithmicpose]}
\algnewcommand\algorithmicmapping{\textbf{Marker mapping:}}
\algnewcommand\Mapping{\item[\algorithmicmapping]}
\algnewcommand\algorithmiccam{\textbf{Camera localisation:}}
\algnewcommand\CamLocalisation{\item[\algorithmiccam]}
\algnewcommand\algorithmicoutput{\textbf{Output:}}
\algnewcommand\OUTPUT{\item[\algorithmicoutput]}

\makeatletter
\newcommand{\oset}[2]{%
	{\mathop{#2}\limits^{\vbox to -.50\ex@{\kern-\tw@\ex@
				\hbox{\scriptsize #1}\vss}}}}
\makeatother

\newcommand{\cG}{\mathcal{G}}

\newcommand{\cA}{\mathcal{A}}

\newcommand{\cB}{\mathcal{B}}

\newcommand{\cC}{\mathcal{C}}

\newcommand{\bc}{\mathbf{c}}

\newcommand{\cE}{\mathcal{E}}

\newcommand{\cS}{\mathcal{S}}

\newcommand{\bu}{\mathbf{u}}

\newcommand{\bp}{\mathbf{p}}

\newcommand{\bq}{\mathbf{q}}
\newcommand{\bR}{\mathbf{R}}

\newcommand{\bt}{\mathbf{t}}

\newcommand{\cZ}{\mathcal{Z}}

\newcommand{\bK}{\mathbf{K}}

\newcommand{\cM}{\mathcal{M}}

\newcommand{\cV}{\mathcal{V}}

\IEEEoverridecommandlockouts                              

\title{\LARGE \bf
Resolving Marker Pose Ambiguity by Robust Rotation Averaging\\with Clique Constraints*
}

\author{Shin-Fang Ch'ng$^{1}$, Naoya Sogi$^{2}$, Pulak Purkait$^{1}$, Tat-Jun Chin$^{1}$ and Kazuhiro Fukui$^{2}$
\thanks{*This work was supported by the ARC Centre of Excellence on Robotic Vision CE140100016 and the Mawson Lakes Fellowship Program.}
\thanks{$^{1}$School of Computer Science, The University of Adelaide, Australia.}%
\thanks{$^{2}$Department of Computer Science, University of Tsukuba, Japan.}%
}

\begin{document}
\maketitle
\thispagestyle{empty}
\pagestyle{empty}
\bstctlcite{IEEEexample:BSTControl}

\begin{abstract}
Planar markers are useful in robotics and computer vision for mapping and localisation. Given a detected marker in an image, a frequent task is to estimate the 6DOF pose of the marker relative to the camera, which is an instance of planar pose estimation (PPE). Although there are mature techniques, PPE suffers from a fundamental ambiguity problem, in that there can be more than one plausible pose solutions for a PPE instance. Especially when localisation of the marker corners is noisy, it is often difficult to disambiguate the pose solutions based on reprojection error alone. Previous methods choose between the possible solutions using a heuristic criteria, or simply ignore ambiguous markers.

We propose to resolve the ambiguities by examining the consistencies of a set of markers across multiple views. Our specific contributions include a novel rotation averaging formulation that incorporates long-range dependencies between possible marker orientation solutions that arise from PPE ambiguities. We analyse the combinatorial complexity of the problem, and develop a novel lifted algorithm to effectively resolve marker pose ambiguities, without discarding any marker observations. Results on real and synthetic data show that our method is able to handle highly ambiguous inputs, and provides more accurate and/or complete marker-based mapping and localisation.
\end{abstract}

\section{Introduction}\label{intro}

In many robotic vision pipelines, fiducial markers are often employed to simplify feature extraction. In particular, planar markers~{\cite{wang2016apriltag,fiala2004artag, fiala2005artag,garrido2014automatic,romero2018speeded, hu2019deep}}, which are designed to be easily detected and associated across images, find extensive use in laboratory and commercial settings (factories, warehouses, mines, etc.). In applications that perform planar marker-based SfM or SLAM~\cite{shaya2012self,munoz2018mapping,degol2018improved,munoz2019spm}, there is a basic need to estimate the 6DOF pose of an observed marker relative to the camera coordinate frame. This is often solved as a special case of planar pose estimation (PPE), which functions by determining the relative pose between a plane of known dimensions and its projection onto the image~\cite{oberkampf1993iterative, schweighofer2006robust, collins2014infinitesimal}.


While in theory 6DOF pose can be determined uniquely from four non-colinear but co-planar points, the situation is less clear in non-ideal conditions where perspective effects are not apparent, e.g., when the imaged marker is small or the marker is at a distance which is significantly larger than the focal length. In such conditions there is a two-fold \textit{rotational ambiguity} that corresponds to an unknown reflection of the plane about the z-axis of the camera~\cite{oberkampf1993iterative, schweighofer2006robust, collins2014infinitesimal}. For one observed planar marker (specifically its four corners), state-of-the-art PPE methods~\cite{schweighofer2006robust,collins2014infinitesimal} may return two physically plausible pose solutions, with one of them being the correct one (i.e., the one closer to the ground truth pose).

\begin{figure}[t]\centering
	\subfigure[]{\includegraphics[height=9em,width=0.22\textwidth]{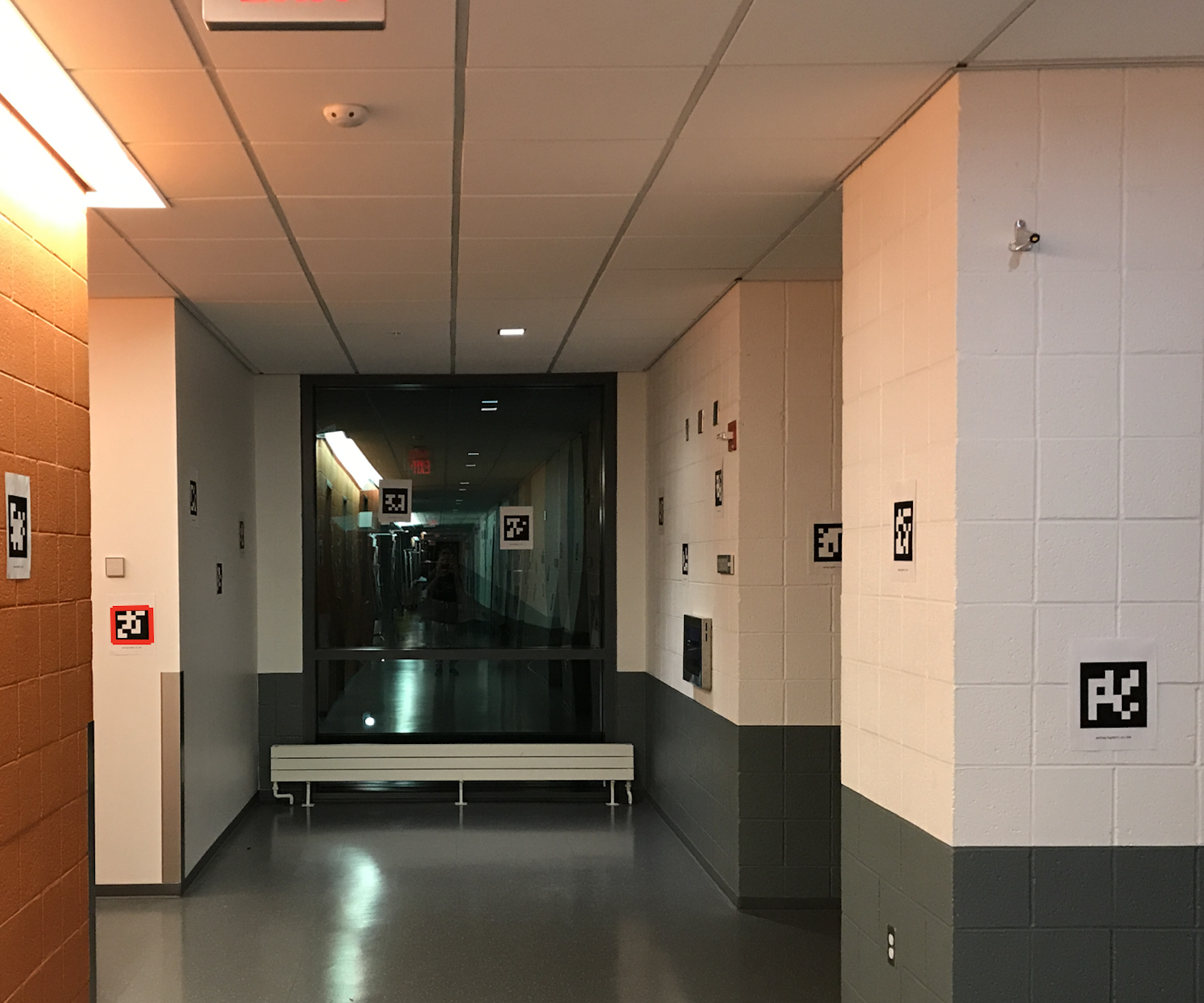}}
	\subfigure[]{\includegraphics[height=9em,width=0.25\textwidth]{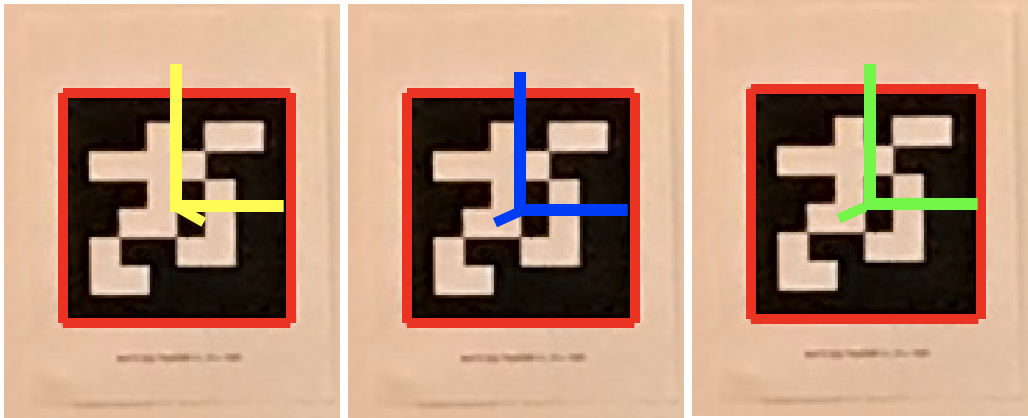}}
	\caption{(a) A detected marker with bounding box from a frame in the dataset of~{\cite{degol2018improved}}. (b) The two poses $\bp'$ (\textit{yellow}) and $\bp''$ (\textit{blue}) returned by PPE~{\cite{collins2014infinitesimal}} have reprojection errors $0.00011$ and $0.00013$ resp. Though $\bp'$ has the lower error, it is an incorrect pose, cf.~the ground truth pose (\textit{green}).}
	\vspace{-0.5cm}
	\label{fig:example_real_world}
\end{figure}

Fig.~\ref{fig:example_real_world} shows an example from the dataset of~{\cite{degol2018improved}}. Note that the two solutions returned by PPE can be very different, thus it is unwise to arbitrarily choose one of the two poses, or take the midpoint of the two solutions as the pose estimate.

A common way to disambiguate the two returned poses $\bp'$ and $\bp''$ is to compute the reprojection error of each pose
\begin{align}\label{eq:reprojection_err}
r(\bp) = \sum_{k=1}^{4} \left\| f( \bK, \bc_k, \bp ) - \bu_k \right\|^2_2,\, \bp\in\{\bp', \bp''\}
\end{align}
where $\{ \bc_k \}^{4}_{k=1}$ and $\{ \bu_k \}^{4}_{k=1}$ are the reference 3D position and 2D observation of the 4 corners of the detected marker, $\bK$ is the camera intrinsic parameter and $f(\bK,\bc,\bp)$ projects $\bc$ onto the image with camera pose $\bp$. The PPE pose with the lower reprojection error is then selected.

However, comparing reprojection errors is not foolproof~\cite{wu2012stable,munoz2019spm}, for if the corner localisation is noisy, $r(\bp')$ and $r(\bp'')$ can be very close. In fact, the correct solution can have the higher reprojection error; see Fig.~\ref{fig:example_real_world}.

\begin{figure}[t]\centering
	\subfigure[Histogram of error ratio~\eqref{eq:resratio}.]{\includegraphics[width=0.23\textwidth]{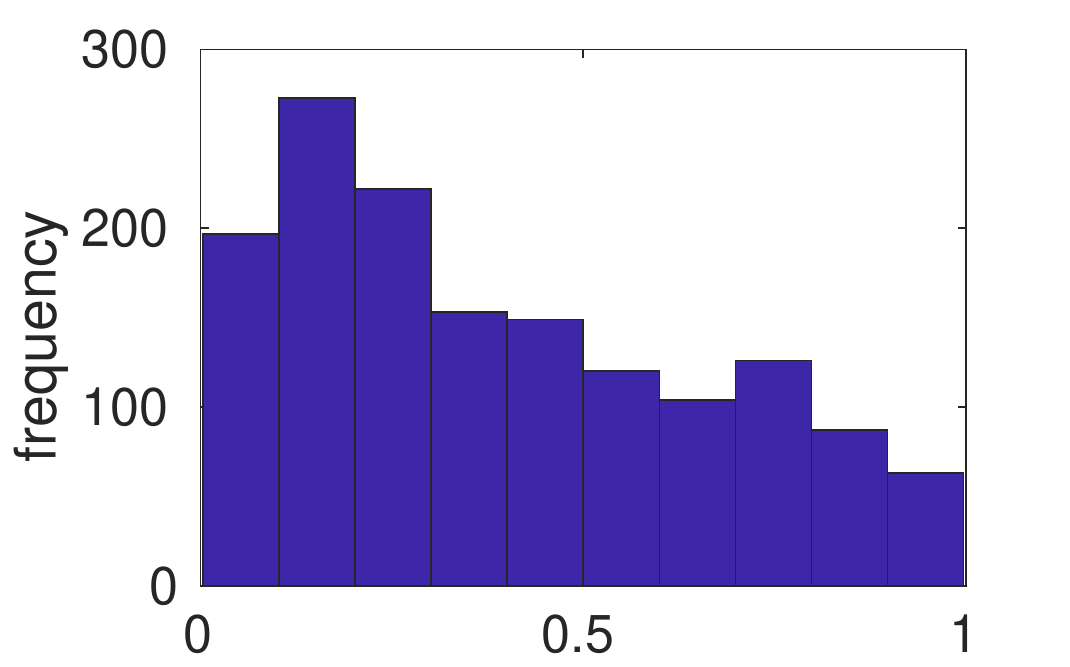}\label{fig:resratio}}	
	\subfigure[Histogram of weight ratio~\eqref{eq:weightratio}.]{\includegraphics[width=0.23\textwidth]{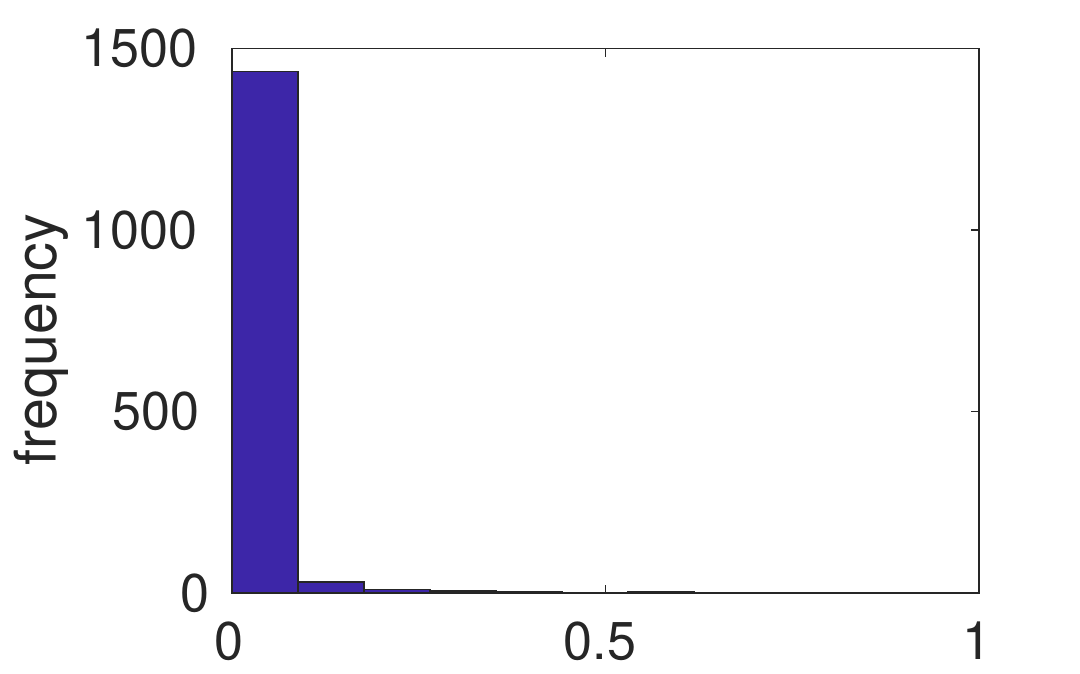}\label{fig:weightratio}}
	\caption{Histogram of reprojection error ratio~\eqref{eq:resratio} and weight ratio~\eqref{eq:weightratio} from proposed method (Sec.~\ref{sec:weighted_maximal_clique_formulation_and_algorithm}) for all markers detected in \textbf{Hotel2}~\cite{dai2017scannet}.}
	\label{fig:statistical_analysis}
\end{figure}

In practice, marker pose ambiguity occurs regularly~\cite{munoz2018mapping}. Fig.~\ref{fig:resratio} is the histogram of the reprojection error ratio
\begin{align}\label{eq:resratio}
\frac{\min\left[r(\bp'),r(\bp'')\right]}{\max\left[r(\bp'),r(\bp'')\right]}
\end{align}
of the PPE-derived poses for all the markers detected in sequence Hotel2\textbf{(H2)} from~\cite{dai2017scannet}. About 25\% of the PPE solutions are considered ambiguous (ratio value $\ge 0.6$~\cite{munoz2018mapping}).


While current theory and algorithms for PPE~\cite{schweighofer2006robust,collins2014infinitesimal} have characterised the ambiguity issue and are able to compute all physically plausible solutions stably, using the PPE outputs under ambiguity, particularly in marker-based SfM or SLAM pipelines, remains a fundamental challenge. In the following, we further survey efforts to deal with marker pose ambiguity, before outlining the proposed solution.

\subsection{Related work}\label{sec:related}

Tanaka et al.~\cite{tanaka2014solution,tanaka2017solving} modified the conventional planar marker design to directly incorporate orientation information. They attach two one-dimensional moire patterns onto the marker to obtain appearance variation for pose disambiguation, as well as lenticular lenses that introduce 3D deviations to the marker surface. Though this largely alleviates the ambiguity problem, the marker fabrication is non-trivial. 

For planar target camera tracking, a filtering method with a well-tuned camera motion model~\cite{wu2012stable,uematsu2007improvement} can be exploited to disambiguate the marker poses. However, this assumes temporal continuity in the images, which may not be valid in SfM with wide baseline images; moreover, there are no mature \emph{filtering methods} for marker SLAM. Jin et al.~\cite{jin2017sensor} showed improved marker pose estimation accuracy by fusing depth information. However, this requires an RGBD camera. 

Marker-based SfM/SLAM is an active research area~\cite{shaya2012self,neunert2016open, munoz2018mapping,degol2018improved,munoz2019spm}. Marker ambiguity is not dealt with explicitly in~\cite{neunert2016open,shaya2012self,degol2018improved}, though~\cite{degol2018improved} combined feature-based SfM with marker-based SfM. Munoz-Salinas et al.~applied the ratio test of~\cite{collins2014infinitesimal} in their marker-based SfM~\cite{munoz2018mapping} and SLAM pipeline~\cite{munoz2019spm}. Basically, if the ratio~\eqref{eq:resratio} is below a threshold (default is 0.6~\cite{munoz2018mapping}), the PPE solution with the lower reprojection error is used in subsequent SfM/SLAM processing; else, the marker detection is discarded. A weakness of this approach is the sensitivity to the threshold. If it is too low, many marker detections will be excluded, leading to data wastage or even SfM/SLAM failure. On the other hand, a high threshold risks using bad marker poses (recall that the pose with the lower reprojection error may not be the correct one) for SfM/SLAM. Sec.~\ref{sec:results} will demonstrate this shortcoming.

\subsection{Our contributions}

Unlike previous works that have used a \emph{per-marker} approach to resolve marker ambiguity, we exploit \emph{multi-view constraints} for disambiguation. From the input marker detections, we first construct a \emph{multigraph} of relative rotation measurements, which incorporates all PPE pose ambiguities. Then, we formulate a novel rotation averaging problem with clique constraints that respects \emph{consistency} (details later) between subsets of relative pose measurements. We examine the combinatorial complexity of the new problem, and develop a lifted optimisation method to efficiently solve it. Then, a series of small maximal weighted clique problems are solved to make the final pose selections.  Our method allows all valid PPE pose combinations to be examined, and leads to more accurate and/or complete marker-based SfM.

\section{Problem formulation}

Consider $T$ input images $\{ I_t \}^{T}_{t=1}$ that observed a set of $N$ markers $\{ \cM_i \}^{N}_{i=1}$ of known sizes  in a static scene. We assume calibrated cameras. A standard marker detection and id algorithm~\cite{opencv_library} is applied to each image. Denote by
\begin{align}
\cA^{t}=\{i \in \{1,\dots,N \} \mid \cM_i~\text{was detected in}~I_t \}
\end{align}
as the set of markers detected in $I_t$. Using a PPE technique~\cite{schweighofer2006robust, collins2014infinitesimal} on the corners of $\cM_i$ detected in $I_t$, the \emph{marker-to-camera (M2C)} relative pose of $\cM_i$ to $I_t$ is computed, which can potentially yield two solutions
\begin{align}\label{eq:relposes}
\left\{ \tilde{\bp}^{(t,0)}_i, \tilde{\bp}^{(t,1)}_i \right\} = \left\{ \tilde{\bp}^{(t,a)}_i \right\}_{a = 0,1}.
\end{align}
Without loss of generality, we assume that each marker observation has exactly two relative pose solutions. Note that the pose ambiguity is due to orientation ambiguity, thus the translation component is the same, i.e.,
\begin{align}
\tilde{\bp}^{(t,0)}_i = \left(\tilde{\bt}_i^{(t)}, \tilde{\bR}_i^{(t,0)}\right), \;\;\;\; \tilde{\bp}^{(t,1)}_i = \left(\tilde{\bt}_i^{(t)}, \tilde{\bR}_i^{(t,1)}\right).
\end{align}

Given the set of all M2C relative pose measurements
\begin{align}\label{eq:allrelposes}
\bigcup\limits_{t=1}^{T} \bigcup\limits_{i \in \cA^t} \left\{ \tilde{\bp}^{(t,a)}_i \right\}_{a=0,1},
\end{align}
our overall aim is SfM, i.e., find the absolute poses of the markers $\{ \bp_i \}^{N}_{i=1}$ and cameras $\{ \bq_{t }\}^{T}_{t=1}$. To do so, pose ambiguity must be resolved, i.e., for each $(i,t)$ such that $i \in \cA^t$, choose \emph{either} $\tilde{\bp}^{(t,0)}_i$ \emph{or} $\tilde{\bp}^{(t,1)}_i$ for SfM computations.

Previous pipelines~\cite{munoz2018mapping,munoz2019spm} make the choice using per-marker heuristics, or discard the marker observation. This ``preprocessing" yields the reduced measurement set
\begin{align}\label{eq:allrelposesresolved}
\bigcup\limits_{t=1}^{T} \bigcup\limits_{i \in \cB^t} \left\{\tilde{\bp}^{(t)}_i \right\},
\end{align}
where each $\tilde{\bp}_i^{(t)}$ is \emph{either} $\tilde{\bp}_i^{(t,0)}$ \emph{or} $\tilde{\bp}_i^{(t,1)}$, and $\cB^t \subseteq \cA^t$. The reduced measurement set is then subjected to the rest of the SfM/SLAM pipeline. Our new method exploits multi-view consistency to disambiguate the PPE marker poses in a way that avoids premature decisions; details as follows.

\section{Multigraph with rotational ambiguity}\label{sec:marker_viewgraph}

\begin{figure*}[t]\centering
	\subfigure[]{\includegraphics[height=4cm,width=0.10\textwidth]{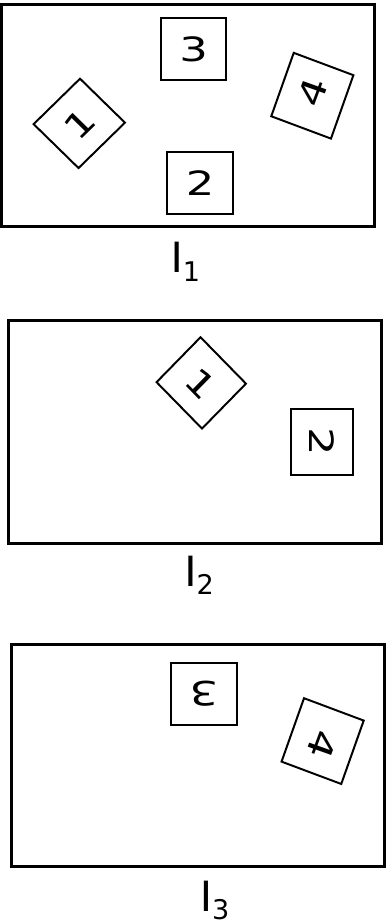}}
	\hfill
	\subfigure[]{\includegraphics[width=0.4\textwidth]{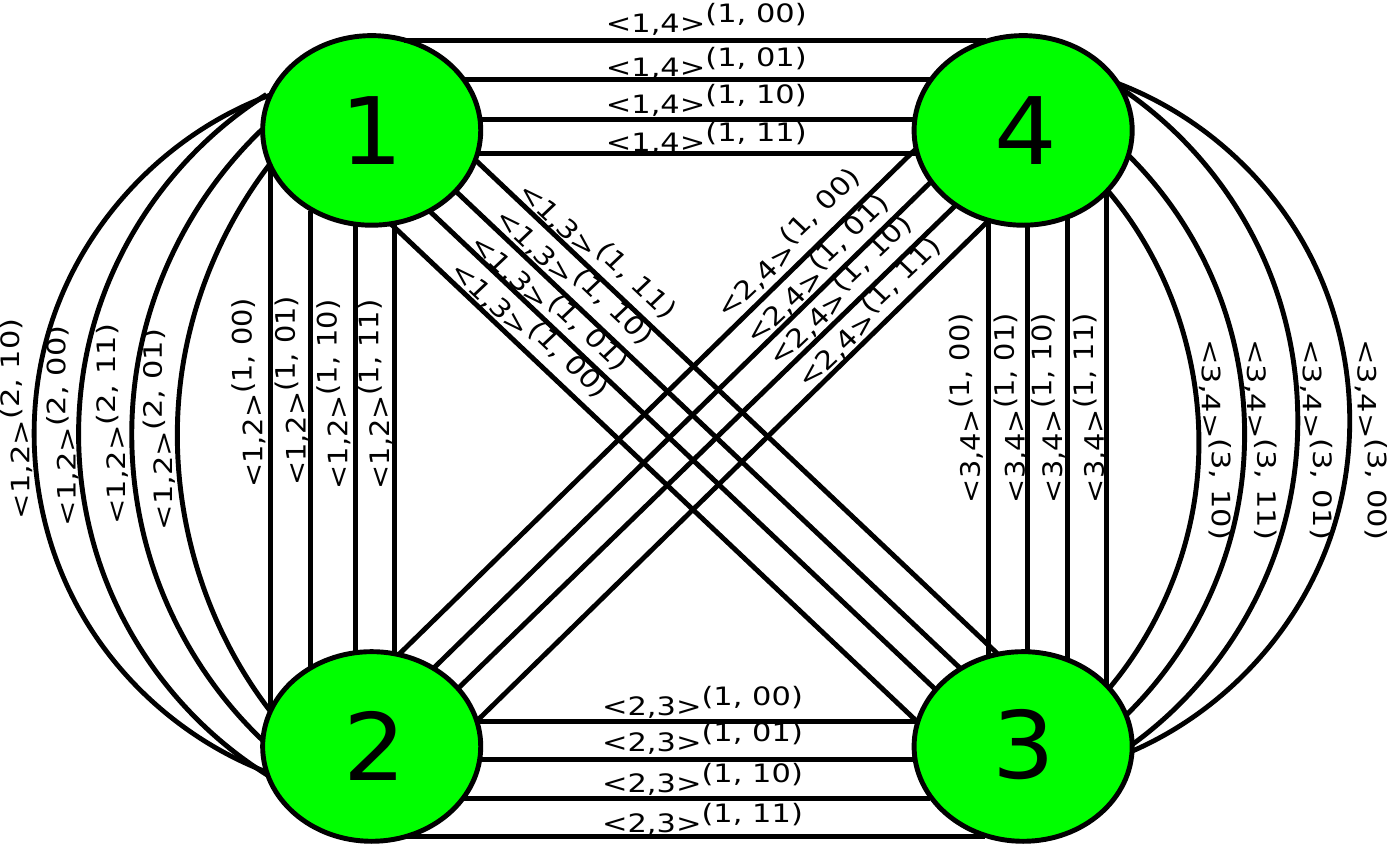}}
	\hfill
	\subfigure[]{\includegraphics[width=0.4\textwidth]{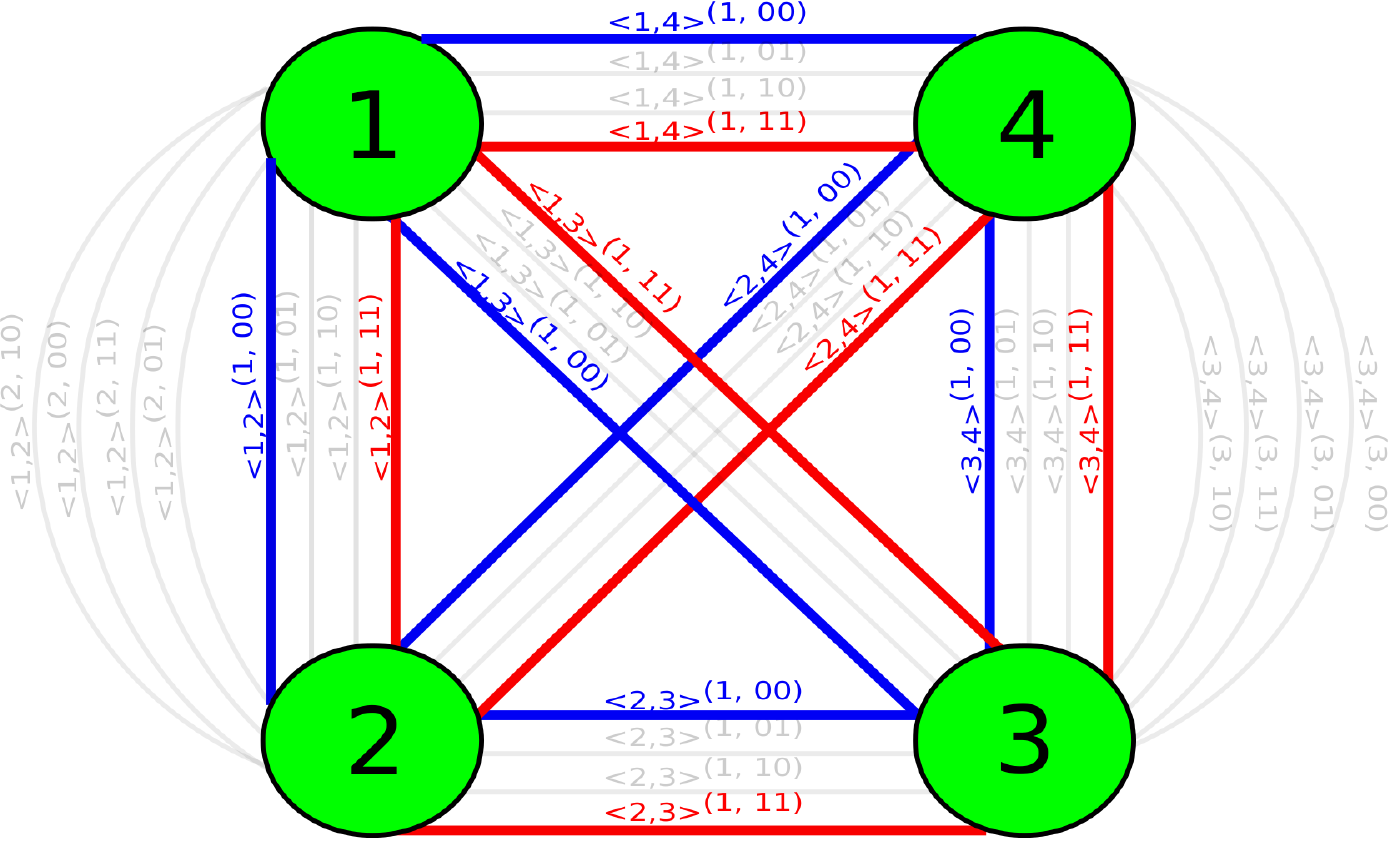}}
	\caption{Multigraph and consistent cliques. (a) The scene has 4 markers $\{ \cM_1, \cM_2, \cM_3, \cM_4 \}$ captured in 3 images $\{I_1, I_2, I_3\}$. All markers were detected in $I_1$, while only a subset was detected in $I_2$ and $I_3$.~\label{subfig:simple_example} (b) Multigraph with the edges labelled following~\eqref{eq:alledges}. Since $\cM_{1}$ and $\cM_{2}$ were covisible in $I_1$ and $I_2$, there are 8 edges connecting vertices $1$ and $2$ (similarly, $\cM_3$ and $\cM_4$ in $I_1$ and $I_3$). (c) Two consistent cliques (red and blue) for image $I_1$. }
	\vspace{-0.5cm}
	\label{fig:consistent_cliques}
\end{figure*}

Since the ambiguity lies in the orientations, it is natural to model the ambiguity using only the M2C relative rotations
\begin{align}\label{eq:allrelrots}
\bigcup\limits_{t=1}^{T} \bigcup\limits_{i \in \cA^t} \left\{ \tilde{\bR}^{(t,a)}_i \right\}_{a = 0,1}.
\end{align}
To this end, we construct a multigraph $\cG =\{\cV, \cE\}$, where the vertices $\cV$ is the set of markers $\{1,\dots,N \}$, and the edges $\cE$ indicate covisibility between the markers. More specifically, if $\cM_i$ and $\cM_j$ are detected in $I_t$, four edges
\begin{align}\label{eq:alledges}
\langle i,j \rangle^{(t,00)}, \;\; \langle i,j \rangle^{(t,01)}, \;\; \langle i,j \rangle^{(t,10)}, \;\; \langle i,j \rangle^{(t,11)}
\end{align}
connect vertices $i$ and $j$ in $\cG$; assuming $i<j$, the edges correspond to the \emph{marker-to-marker (M2M)} relative rotations
\begin{equation}\label{eq:relative_pose_between_pairwise_markers}
\begin{aligned}
\tilde{\bR}_{i,j}^{(t,00)} &=  (\tilde{\bR}_{j}^{(t,0)})^{T}\tilde{\bR}_{i}^{(t,0)}, \\
\tilde{\bR}_{i,j}^{(t,01)} &=  (\tilde{\bR}_{j}^{(t,1)})^{T}\tilde{\bR}_{i}^{(t,0)},\\
\tilde{\bR}_{i,j}^{(t,10)} &=  (\tilde{\bR}_{j}^{(t,0)})^{T}\tilde{\bR}_{i}^{(t,1)}, \\
\tilde{\bR}_{i,j}^{(t,11)} &=  (\tilde{\bR}_{j}^{(t,1)})^{T}\tilde{\bR}_{i}^{(t,1)}.
\end{aligned}
\end{equation}
Fig.~\ref{fig:consistent_cliques} shows an example. Since multiple edges connect two vertices, $\cG$ is a multigraph. We summarise~\eqref{eq:alledges} and~\eqref{eq:relative_pose_between_pairwise_markers} as
\begin{align}
\left\{ \langle i,j \rangle^{(t,ab)}  \right\}_{ab = 00,01,10,11}, \;\; \left\{ \tilde{\bR}_{i,j}^{(t,ab)}  \right\}_{ab = 00,01,10,11},
\end{align}
where $ab$ is a bit string composed of two binary indicators $a,b \in \{0,1\}$. The edges in $\cG$ are undirected; if $i < j$, the edge $\langle j,i \rangle^{(t,ab)}$ has the associated M2M relative rotation
\begin{align}
\tilde{\bR}_{j,i}^{(t,ab)} = (\tilde{\bR}_{j}^{(t,a)})^{T}\tilde{\bR}_{i}^{(t,b)}.
\end{align}
Thus, in our notation
\begin{align}
\langle i,j \rangle^{(t,ab)} = \langle j,i \rangle^{(t,ba)} \ne \langle j,i \rangle^{(t,ab)}.
\end{align}
The set of all edges $\cE$ (without repetitions) is thus
\begin{align}
\cE = \bigcup\limits_{t=1}^{T} \bigcup\limits_{\substack{i,j \in \cA^t\\ i < j}} \left\{ \langle i,j \rangle^{(t,ab)}  \right\}_{ab = 00,01,10,11}.
\end{align}
Similarly, the set of unique M2M relative rotations is
\begin{align}
\bigcup\limits_{t=1}^{T} \bigcup\limits_{\substack{i,j \in \cA^t\\ i<j}} \left\{ \tilde{\bR}_{i,j}^{(t,ab)}  \right\}_{ab = 00,01,10,11}.
\end{align}
The existence of four M2M relative rotations per $\langle i,j \rangle$ pair is a direct consequence of ambiguity in marker pose estimation, and the bit string $ab$ selects a particular combination of M2C relative rotations to derive the M2M relative rotation.


Note that our multigraph construction method is a significant extension of that in~\cite{munoz2018mapping}, in that our multigraph incorporates all ambiguous marker poses, whereas~\cite{munoz2018mapping} generates $\cG$ from the preprocessed data~\eqref{eq:allrelposesresolved} with no ambiguities.

\subsection{Consistent cliques}\label{sec:complexity}

We assume that the multigraph $\cG$ is connected, i.e., there is a path that connects every pair of vertices (markers) in $\cG$.  

\begin{defn}(Consistent clique)\label{def:clique} Given multigraph $\cG = \{\cV,\cE \}$ as defined above, a consistent clique for image $I_t$ is a fully connected subgraph $\cC = \{ \cV', \cE' \}$ such that
\begin{itemize}
\item $\cV' = \cA^t  \subseteq \cV$;
\item Every two vertices $i,j \in \cV'$ are connected by \textbf{exactly} one edge $\langle i,j \rangle^{(t,ab)}$, where
$ab$ is one of $\{00,01,10,11\}$.
\item For every two vertices $j,k \in \cV'$ that are connected to vertex $i$, the associated edges $\langle i,j \rangle^{(t,ab)}$ and $\langle i,k \rangle^{(t,cd)}$ satisfy the condition $a = c$.
\end{itemize}	
\end{defn}
Fig. \ref{fig:consistent_cliques} provides examples. Intuitively, a consistent clique $\cC$ for image $I_t$ corresponds to a set of M2M relative rotations that are composed using a constant selection of one of the two M2C relative poses for each marker detected in $I_t$.



Since there are multiple valid combinations of constant M2C relative pose selections, there are multiple consistent cliques for an image. Assuming that $V$ markers are detected in each image, there are $\mathcal{O}(2^V)$ number of consistent cliques per image. For $T$ images, there are thus $\mathcal{O}(2^{VT})$ unique combinations of consistent cliques across the images.

\section{Disambiguation with rotation averaging}\label{sec:proposed}

Based on the multigraph, our technique resolves the ambiguities by first solving a novel rotation averaging formulation, then - based on the averaging results - building and solving a maximum weighted clique problem. The key outcome of this step is marker pose disambiguation;  Sec.~\ref{sec:overall} will incorporate this step into a marker-based SfM pipeline.

\subsection{Rotation averaging with clique constraints} \label{subsec:extending_ra}

While standard rotation averaging is defined over a graph of relative rotations~\cite{hartley2013rotation,chatterjee2013efficient}, extending the formulation to a multigraph of relative rotations is straightforward, and existing algorithms (we used~\cite{chatterjee2013efficient}) can be applied with minor adjustments. Let $\{ \bR_i \}_{i=1}^N$ be the absolute rotations of the markers. A rotation averaging problem over multigraph $\cG$ is
\begin{align} \label{eq:extended_rot_avg_multigraph}
\min_{\{\bR_{i}\}^{N}_{i=1}} \sum_{t=1}^{T} \sum_{\substack{i,j \in \cA^t\\ i < j}}  \sum_{a,b \in \{0,1\}} \hspace{-1em} \rho \left(\left\|{\tilde \bR}_{i,j}^{(t,ab)} - \bR_{j}\bR_{i}^{T}\right\|_F\right),
\end{align}
where $\rho$ is a robust norm. The motivation behind~\eqref{eq:extended_rot_avg_multigraph} is to attempt to identify the incorrect poses from PPE as the contributors to outlying measurements in the averaging task.

However, our tests (Sec.~\ref{sec:results}) suggest that this approach is ineffective for disambiguation, most probably because \eqref{eq:extended_rot_avg_multigraph} does not enforce clique consistency (Def.~\ref{def:clique}). Thus, error terms that are regarded as inliers could correspond to choosing \emph{both} PPE poses for the same marker detection.

To enforce clique consistency into rotation averaging, we introduce a set of binary indicator variables
\begin{align}
\cS = \bigcup_{t=1}^T \{ s^t_i \in \{ 0,1\} \mid i \in \cA^t\},
\end{align}
where the setting $s^t_i = 0$ implies selecting M2C relative rotation~$\tilde{\bR}^{(t,0)}_i$ the detection of $\cM_i$ in $I_t$, while $s^t_i = 1$ implies selecting $\tilde{\bR}^{(t,1)}_i$. We then formulate the clique-constrained rotation averaging problem
\begin{equation} \label{eq:lifted_switchable_exclusion_rot_avg_multigraph_compact_form}
\begin{aligned}
\min_{\substack{\{\bR_{i}\}^{N}_{i=1}, \cS}} \sum_{t=1}^{T} \sum_{\substack{i,j \in \cA^t\\ i < j}}
s_i^t  s_j^t \left\|{\tilde \bR}_{i,j}^{(t,11)} - \bR_{j}\bR_{i}^{T}\right\|_F& + \\
 s_i^t  (1-s_j^t) \left\|{\tilde \bR}_{i,j}^{(t,10)} - \bR_{j}\bR_{i}^{T}\right\|_F& + \\
(1-s_i^t)  s_j^t \left\|{\tilde \bR}_{i,j}^{(t,01)} - \bR_{j}\bR_{i}^{T}\right\|_F& + \\
(1- s_i^t)(1- s_j^t) \left\|{\tilde \bR}_{i,j}^{(t,00)} - \bR_{j}\bR_{i}^{T}\right\|_F&.
\end{aligned}
\end{equation}
Intuitively, $\cS$ selects the M2C relative rotations to compose the M2M relative rotations in a consistent way. Searching over $\cS$ thus allows different consistent cliques in all images to be examined. Finally, since $\{ \bR_i \}^{N}_{i=1}$ are shared across images, multi-view consistency is exploited to choose the best combinations of the PPE relative rotations.


\subsection{Efficient algorithm using lifting approach}\label{subsec:lifting}

A naive method to solve~\eqref{eq:lifted_switchable_exclusion_rot_avg_multigraph_compact_form} is to enumerate $\cS$, and for each $\cS$ instantiation, collect the non-zero terms in~\eqref{eq:lifted_switchable_exclusion_rot_avg_multigraph_compact_form} and solve the resulting rotation averaging problem. Then, return the $\cS$ with the lowest optimised error as the disambiguation decision. Since there are $\mathcal{O}(2^{VT})$ possible instantiations of $\cS$ (assuming $V$ markers seen per image), this is infeasible.

To enable an efficient algorithm for~\eqref{eq:lifted_switchable_exclusion_rot_avg_multigraph_compact_form}, we apply the lifting approach~\cite{sunderhauf2012towards}. First, we relax the indicator variables $s^t_i \in [0,1]$ and replace them in~\eqref{eq:lifted_switchable_exclusion_rot_avg_multigraph_compact_form} with a sigmoid function
\begin{align}
\Phi(s)=1/(1+e^{-s}),
\end{align}
which yields the ``smoothed" version of~\eqref{eq:lifted_switchable_exclusion_rot_avg_multigraph_compact_form}
\begin{equation}\label{eq:lifted_algorithm}
\begin{aligned}
\min_{\substack{\{\bR_{i}\},\cS }} \sum_{t=1}^{T} \sum_{\substack{i,j \in \cA^t\\ i < j}}
\Phi(s_i^t) \Phi(s_j^t) \left\|{\tilde \bR}_{i,j}^{(t,11)} - \bR_{j}\bR_{i}^{T}\right\|_F& + \\
\Phi(s_i^t) (1-\Phi(s_j^t)) \left\|{\tilde \bR}_{i,j}^{(t,10)} - \bR_{j}\bR_{i}^{T}\right\|_F& + \\
(1-\Phi(s_i^t))  \Phi(s_j^t) \left\|{\tilde \bR}_{i,j}^{(t,01)} - \bR_{j}\bR_{i}^{T}\right\|_F& + \\
(1- \Phi(s_i^t))(1- \Phi(s_j^t)) \left\|{\tilde \bR}_{i,j}^{(t,00)} - \bR_{j}\bR_{i}^{T}\right\|_F&.
\end{aligned}
\end{equation}
Intuitively, the contribution of an error term in~\eqref{eq:lifted_algorithm} is now weighted according to correctness of the corresponding M2C relative poses that define the error term.

Problem~\eqref{eq:lifted_algorithm} can be solved using an iterative non-linear optimiser (e.g., \textit{fmincon} in MATLAB). We initialise $\{ \bR_i \}$ via a minimum spanning tree on $\cG$, choosing the M2M relative rotations with the lower combined reprojection errors for chaining, and $\cS$ is set to reflect these choices. As we will show in Sec.~\ref{sec:results}, our method is not biased by such an initialisation, since it is capable of providing more accurate disambiguation than comparing reprojection errors alone.

\subsection{Selecting the marker poses}\label{sec:weighted_maximal_clique_formulation_and_algorithm}

Let $\hat{\cS}$ by the optimised relaxed indicator variables from solving~\eqref{eq:lifted_algorithm}. For the same sequence used in Fig.~\ref{fig:resratio}, we plot in Fig.~\ref{fig:weightratio}  the histogram of the ratios
\begin{align}\label{eq:weightratio}
\frac{\min( \Phi{(\hat{s}^t_i)}, 1- \Phi{(\hat{s}^t_i ))}}{\max( \Phi{(\hat{s}^t_i)}, 1- \Phi{(\hat{s}^t_i )})}
\end{align}
for all $\hat{s}^t_i \in \hat{\cS}$. Similar to~\eqref{eq:resratio}, the ratio~\eqref{eq:weightratio} indicates how ``disambiguable" the PPE poses are for each marker detection (smaller ratios are better), but now based on the value of $\hat{s}^t_i$. Although $\hat{\cS}$ is not discrete, the percentage of marker poses that are still ambiguous is now significantly reduced.


To conclusively select one PPE pose per detected marker, a simple solution would be to threshold each $\hat{s}^t_i \in \hat{\cS}$ with $0.5$; however, we would like to avoid such a per-marker decision. To this end, for each image $I_t$ we construct the multigraph $\cG_t = \{ \cV_t, \cE_t \}$, where $\cV_t = \cA^t$, and
\begin{align}
\cE_t = \left\{ \langle i,j \rangle^{(t,ab)} \mid i,j \in \cA^t, ab \in \{00,01,10,11 \} \right\}.
\end{align}
Note that $\cG_t$ is a submultigraph of $\cG$, and there exist $\mathcal{O}(2^{V})$ consistent cliques in $\cG_t$ (see Sec.~\ref{sec:complexity}). Further, each edge $\langle i,j \rangle^{(t,ab)}$ in $\cG_t$ has the weight
\begin{align}
\hat{w}^{(t,ab)}_{i,j} = \begin{cases} (1-\Phi(\hat{s}^t_i))(1-\Phi(\hat{s}^t_j)) & \text{if}~ab = 00;\\
(1-\Phi(\hat{s}^t_i))\Phi(\hat{s}^t_j) & \text{if}~ab = 01;\\
\Phi(\hat{s}^t_i)(1-\Phi(\hat{s}^t_j)) & \text{if}~ab = 10;\\
\Phi(\hat{s}^t_i)\Phi(\hat{s}^t_j) & \text{if}~ab = 11.
\end{cases}
\end{align}
Given $\cG_t$, define edge indicator variables
\begin{align*}
\cZ_t = \left\{ z_{i,j}^{(t,ab)} \in \{0,1 \} \mid   i,j \in \cA^t, ab \in \{00,01,10,11 \}  \right\}.
\end{align*}
and the maximum weighted clique (MWC) problem
\begin{align}\tag{$MWC_t$}
\begin{aligned}
& \underset{\cZ_t}{\max}
& & \sum_{\substack{i,j \in \cA^t\\ i<j } }\sum_{ab \in \{00,01,10,11 \}} z^{(t,ab)}_{i,j} \hat{w}^{(t,ab)}_{i,j}\\
& \text{s.t.}
& & \{ \langle i,j \rangle^{(t,ab)} \mid z_{i,j}^{(t,ab)} = 1 \}~\text{is consistent}.
\end{aligned}
\end{align}
Basically, the aim of $MWC_t$ is to find a consistent clique in $I_t$ with the largest edge weights. Though MWC is intractable in general~\cite{tomita2003efficient}, each $MWC_t$ instance is small, since the number $V$ of detected markers in $I_t$ is small (usually $V\le 9$).

We use the efficient clique solver of~\cite{eppstein2011listing} on each $MWC_t$. The optimised $\hat{\cZ}_t$ provides a consistent selection of the PPE poses for all markers detected in $I_t$. Specifically, for each $\cM_i$ detected in $I_t$, find a $\hat{z}_{i,j}^{(t,ab)}$ that is nonzero, and set $\tilde{\bp}^{(t)}_i = \tilde{\bp}^{(t,0)}_i$ if $a=0$,  or $\tilde{\bp}^{(t)}_i = \tilde{\bp}^{(t,1)}_i$ otherwise. 

Algorithm~\ref{alg:disambiguation} summarises the proposed method for marker pose disambiguation.

\section{Marker-based SfM pipeline}\label{sec:overall}

To carry out marker-based SfM using our marker pose disambiguation method, we largely follow the pipeline of the state-of-the-art MarkerMapper~\cite{munoz2018mapping}. Briefly, a robust pose graph optimisation is first invoked on the resolved M2C relative poses~\eqref{eq:allrelposesresolved} from Algorithm~\ref{alg:disambiguation} to yield absolute marker poses $\{ \bp_i \}^{N}_{i=1}$ - in our case, the absolute rotation component is initialised using the output $\{ \tilde{\bR}_i \}$ from solving~\eqref{eq:lifted_algorithm}. Then, each camera pose $\bq_t$ is initialised using single pose averaging from the M2C poses, before all marker $\{ \bp_i \}^{N}_{i=1}$ and camera poses $\{ \bq_t \}^{T}_{t=1}$ are refined simultaneously by bundle adjustment on the observed corners of all detected markers. We refer to~\cite{munoz2018mapping} for details of the SfM pipeline.

\begin{algorithm}[t]
	\caption{Method for marker pose disambiguation}
	\label{alg:disambiguation}
	\begin{algorithmic}[1]
		\INPUT{M2C relative poses~\eqref{eq:allrelposes} with PPE ambiguity.}
		\State Construct a multigraph $\cG$ from the input (Sec.~\ref{sec:marker_viewgraph}). 
		\State $\{\hat{\bR}_i \}, \{\hat{s}^t_i \} \leftarrow$ Solve~\eqref{eq:lifted_algorithm} based on $\cG$ (Sec.~\ref{subsec:lifting}).
		\For{$t = 1,\dots,T$}
		\State $\{ \hat{z}_{i,j}^{(t,ab)} \} \leftarrow$ Solve $MWC_t$ from $\{\hat{s}^t_i \}$ (Sec.~\ref{sec:weighted_maximal_clique_formulation_and_algorithm}).
		\State $\{ \tilde{\bp}^{(t)}_i \} \leftarrow$ Based on $\{ \hat{z}_{i,j}^{(t,ab)} \}$, select one of two M2C poses for all markers in $I_t$ (Sec.~\ref{sec:weighted_maximal_clique_formulation_and_algorithm}).
		\EndFor
		\OUTPUT{One M2C relative pose per detected marker.}
	\end{algorithmic}
\end{algorithm}

\section{Results}\label{sec:results}

To assess the efficacy of the proposed marker pose disambiguation technique, we compared the following methods:
\begin{itemize}[leftmargin=1em]
	\item \textbf{Reprojection error (M1)}: For each marker detection, select the PPE solution with the lower reprojection error.
	\item \textbf{Strict ratio test (M2)}: The threshold of $0.1$ is applied on the reprojection error ratio~\eqref{eq:resratio} (see Sec.~\ref{sec:related} for details).
	\item \textbf{Default ratio test (M3)}: The threshold of $0.6$ is applied on the reprojection error ratio (the default setting in~\cite{munoz2018mapping}).
	\item \textbf{Robust rotation averaging and post hoc clique consistency enforcement (M4)}: Solve~(\ref{eq:extended_rot_avg_multigraph}) by IRLS~\cite{chatterjee2013efficient}, then use the IRLS-optimised weights for the error terms as inputs to our M2C pose selection method in Sec.~\ref{sec:weighted_maximal_clique_formulation_and_algorithm}.
	\item \textbf{Proposed method (Ours)}: As described in Sec.~\ref{sec:proposed}.
\end{itemize}
When applying the above disambiguation methods to perform marker-based SfM, we simply used them to preprocess the input marker detections, then execute the rest of the pipeline of MarkerMapper~\cite{munoz2018mapping} (see Sec.~\ref{sec:overall}). All the experiments were conducted on a 3.5GHz CPU and 8GB of RAM.



\subsection{Experiments on hybrid data} \label{subsec:generating hybrid-synthetic data}

\subsubsection{Data generation} \label{subsubsec:perfect_data_generation}

We used the \textbf{ScanNet Dataset}~\cite{dai2017scannet} that contained a number of sequences with ground truth 6DOF camera poses and depth. A test sequence was created from an original sequence by warping a number of ArUco markers~\cite{garrido2014automatic,romero2018speeded} based on known/ground truth M2C relative poses $\bar{\bp}_i^{(t)}$ onto parts of the images that correspond to planar surfaces; see supplementary video~\footnote{https://www.youtube.com/watch?v=LtwavEeCkQ4\&t=\label{footnote:video_link}} for a sample sequence. Using the ground truth camera absolute pose $\bar{\bq}_{t}$, the ground truth marker absolute pose is $\bar\bp_{i} = \bar{\bq}^{-1}_{t} \bar{\bp}_i^{(t)}$.

\begin{table*}[h]
	\centering
	{
		\begin{tabular}{lcc|ccccc|ccccc|ccccc}
			\toprule
			\textbf{Seq} & $N$ & $T$
			& \multicolumn{5}{c}{\textbf{Precision(\%)}} 
			& \multicolumn{5}{c}{\textbf{\# markers mapped}} 
			& \multicolumn{5}{c}{\textbf{\# cameras localised}}\\
			& &  & {M1}& {M2}& {M3} & {M4} & {Ours} 
			& {M1}& {M2}& {M3} & {M4} & {Ours}
			&\multicolumn{1}{c}{M1}
			&\multicolumn{1}{c}{M2}
			&\multicolumn{1}{c}{M3}
			&\multicolumn{1}{c}{M4}
			&\multicolumn{1}{c}{Ours}		\\ 
				\midrule
			\textbf{B}  
			& 3 &31 
			& 94.32 & \textbf{100} & 92.31 & 31.82 & \textbf{100}
			& 3 & {0} & 3 & 3 & 3 
			& 31 & {0} & 31 & 31 & 31 \\
			
			\textbf{H1}  
			& 5 & 41 
			&80.68 & \textbf{100} & 82.61 & 22.16 & \textbf{100}
			& 5 & {0} & 5 & 5 & 5 
			& 41 & {0} & 40 & 41 & 41  \\
			
			\textbf{O1} 
			& 7 & 51
			& 77.08 & \textbf{96.97} & 78.8 & 14.58 & 96.52
			& 7 & 7 & 7 & 7 & 7  
			& 51 & {41} & 51 & 51 & 51 \\ 
			
			\textbf{O2} 
			& 6 & 91
			& 92.64 & \textbf{100} & 98.95 & 37.94 & 99.41
			& 6 & {4} & 6 & 6 & 6  
			& 91 & {46} & 91 & 91 & 91  \\
			
			\textbf{H2} 
			& 14 & 151
			& 93.42 & 98.94 & 97.89 & 48.16 &  \textbf{100}
			& 14 & {13} & 14 & 14 & 14 
			& 151 & {101} & 151 & 151  & 151 \\
			\bottomrule
		\end{tabular}
	}
	\caption{Precision in pose disambiguation on Hybrid Data.}
	\label{tab:synthetic_data_precision_disambiguation}	
\end{table*}

\begin{table*}[t!]
	\centering
	\setlength{\tabcolsep}{5pt} 
	\smallskip\noindent
	{
		\begin{tabular}{l|cc|cc|cc|cc|cc|cc|cc|cc|cc|cc}
			\toprule
			\textbf{Seq} 
			& \multicolumn{10}{c}{\textbf{Average marker pose error ($^{\circ}$, \text{cm})}}
			& \multicolumn{10}{|c}{\textbf{Average camera pose error ($^{\circ}$, \text{cm})}}\\
			&\multicolumn{2}{c}{M1}
			&\multicolumn{2}{c}{M2}
			&\multicolumn{2}{c}{M3}
			&\multicolumn{2}{c}{M4} 
			&\multicolumn{2}{c}{Ours}
			&\multicolumn{2}{|c}{M1}
			&\multicolumn{2}{c}{M2}
			&\multicolumn{2}{c}{M3}
			&\multicolumn{2}{c}{M4}
			&\multicolumn{2}{c}{Ours}\\ 
			\midrule 
			\textbf{B}  
			& 5.4& 11.7 & - & - & 6.3 & 15.0 & 19.0 & 37.5 &\textbf{2.3} & \textbf{2.2}  
			& 7.0 & 15.9 & - & - & 11.9 & 19.5 & 32.0 & 10.0 & \textbf{0.8} & \textbf{2.0}\\
			
			\textbf{H1}  
			& 11.7 & 13.0 & - & - &12.5 & 15.0 & 39.1 & 26.3 &  \textbf{3.3} & \textbf{8.6}
			& 14.8 & 27.5 & - & - & 17.6 & 41.6 & 37.9 & 28.8 & \textbf{5.0} & \textbf{3.2}\\
			
			\textbf{O1} 
			& 26.2 & 30.3 & 15.2 & 8.0 & 25.4 & 29.0 & 55.3 & 120.9 &  \textbf{3.5} & \textbf{4.3} 
			& 17.3 & 69.8 & 7.6 & 16.0 & 19.2 & 69.4 & 85.8 & 49.7 & \textbf{5.7} & \textbf{13.7}\\ 
			
			\textbf{O2} 
			& 8.7 & 6.6 & 4.4 & 4.2 & \textbf{4.1} & \textbf{2.6} & 28.0 &  63.2 & 4.2 & 2.4  
			& 6.2 & 10.5 & \textbf{0.8} & \textbf{2.4} & 17.4 & 4.0 & 41.6 & 40.1 & 1.3 & 3.4 \\			
			
			\textbf{H2} 
			& 4.3 & 5.1 & 7.7 & 3.1 & 5.4 & 5.5 &20.3& 14.2 & \textbf{3.6} & \textbf{4.9} 
			& 4.3 & 3.8 & \textbf{2.2} & \textbf{2.3} & 3.3 & 3.1 & 32.0 & 10.0 &3.4 & 2.4\\			
			\bottomrule
		\end{tabular}
	}
	\caption{SFM Accuracy for different pose disambiguation methods on hybrid data. `-' denotes failed reconstruction.}
	\label{tab:synthetic_data_sfm_accuracy}
\end{table*}

\subsubsection{Marker detection}

Using the steps above, we generated five testing sequences from Bedroom(\textbf{B}), Hotel1(\textbf{H1}), Hotel2(\textbf{H2}), Office1(\textbf{O1}) and Office2(\textbf{O2}). We used~\cite{garrido2014automatic} to detect, identify and localise the corners of each marker in each frame; see Table~\ref{tab:synthetic_data_precision_disambiguation} for the number of frames and unique detected markers in each sequence. Though the markers were synthetically warped into the images, our analysis suggests that corner localisation suffered from errors of 1--7 pixels.




\begin{table*}[t]
	\smallskip\noindent
	\resizebox{\linewidth}{!}
	{
		\centering
		\begin{tabular}{|c|c|c|c|c|}
			\toprule
			\textbf{M1} & \textbf{M3} & \textbf{M4} & \textbf{Ours} & \textbf{FM}  \\
			\cmidrule{1-5}
			&&&&\\ 
			\includegraphics[height=1.5cm,width=0.20\textwidth]{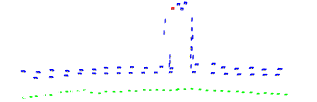}&
			\includegraphics[height=1.5cm,width=0.20\textwidth]{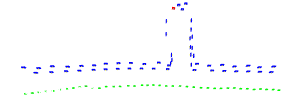} &
			\includegraphics[height=1.5cm,width=0.20\textwidth]{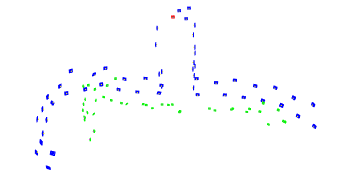} & 
			\includegraphics[height=1.5cm,width=0.20\textwidth]{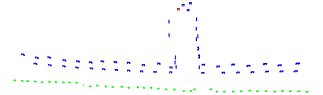}&
			\includegraphics[height=1.5cm,width=0.20\textwidth]{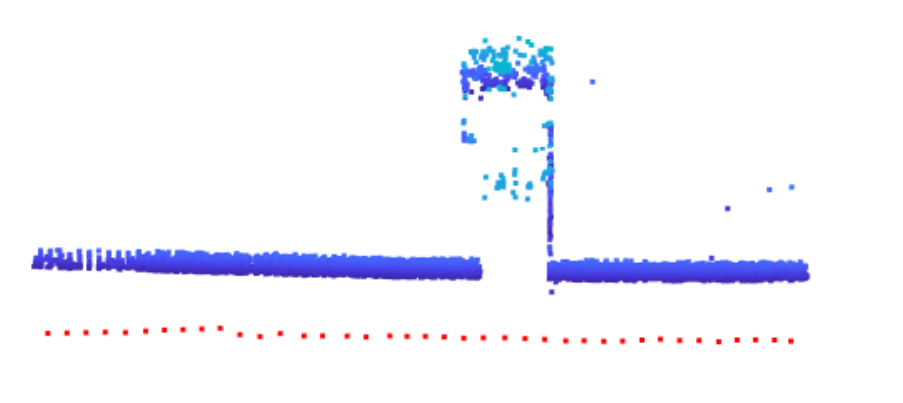}\\   
			\hline  
			&&&&\\ 
			\includegraphics[height=3.5cm,width=0.20\textwidth]{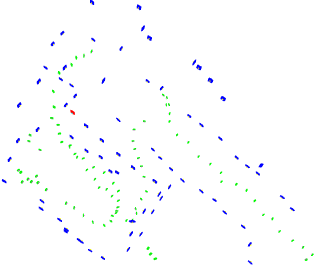}& 
			\includegraphics[height=3.5cm,width=0.20\textwidth]{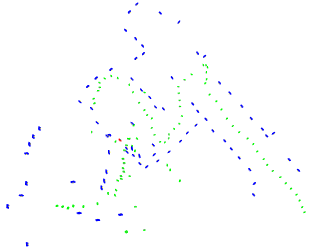}& 		
			\includegraphics[height=3.5cm,width=0.20\textwidth]{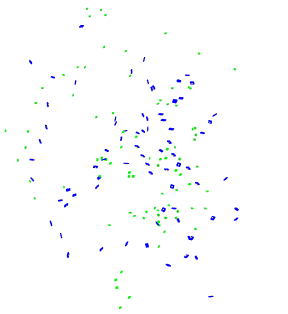}& 
			\includegraphics[height=3.5cm,width=0.20\textwidth]{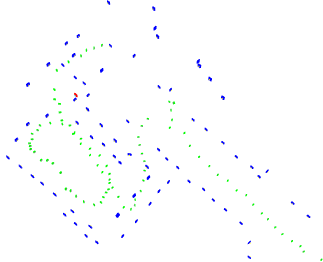} &
			\includegraphics[height=3.5cm,width=0.20\textwidth]{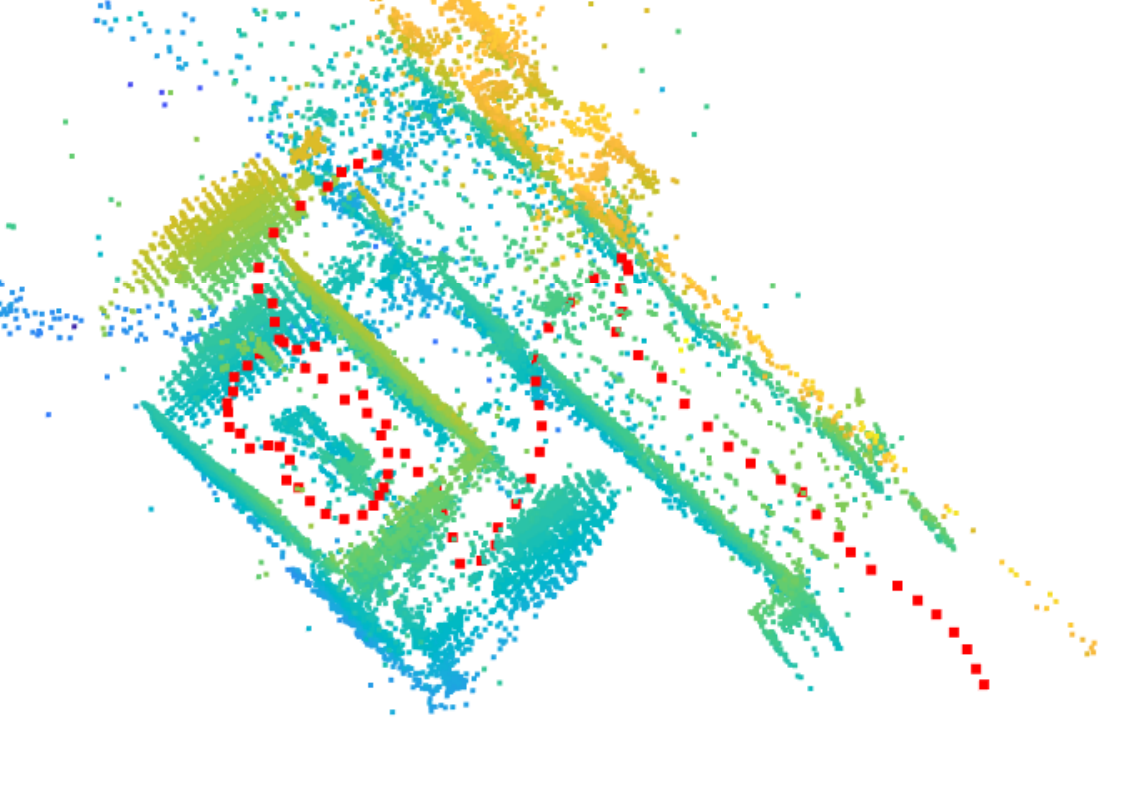} \\ 
			\hline  
			&&&&\\ 
			\includegraphics[width=0.20\textwidth]{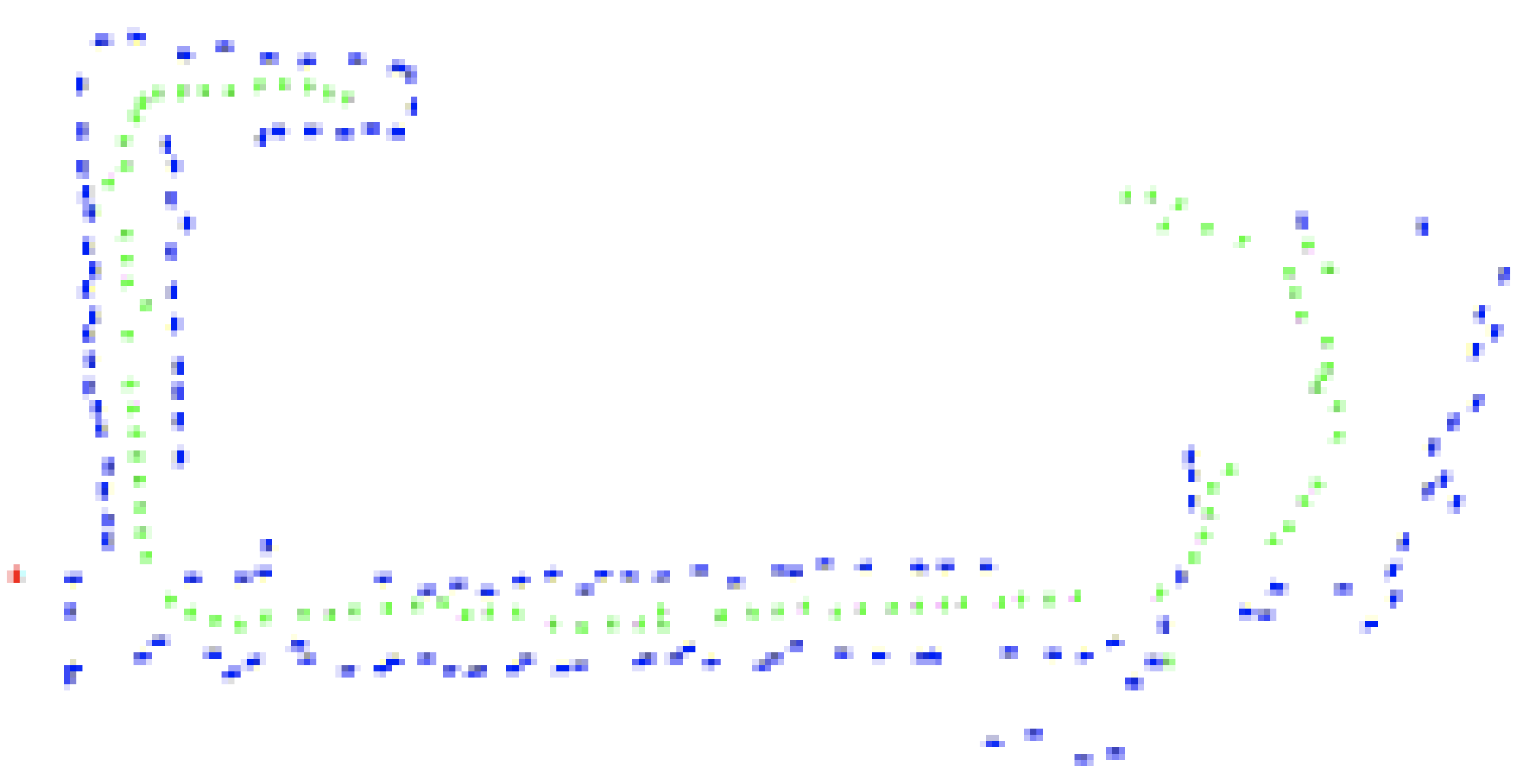}& 
			\includegraphics[width=0.20\textwidth]{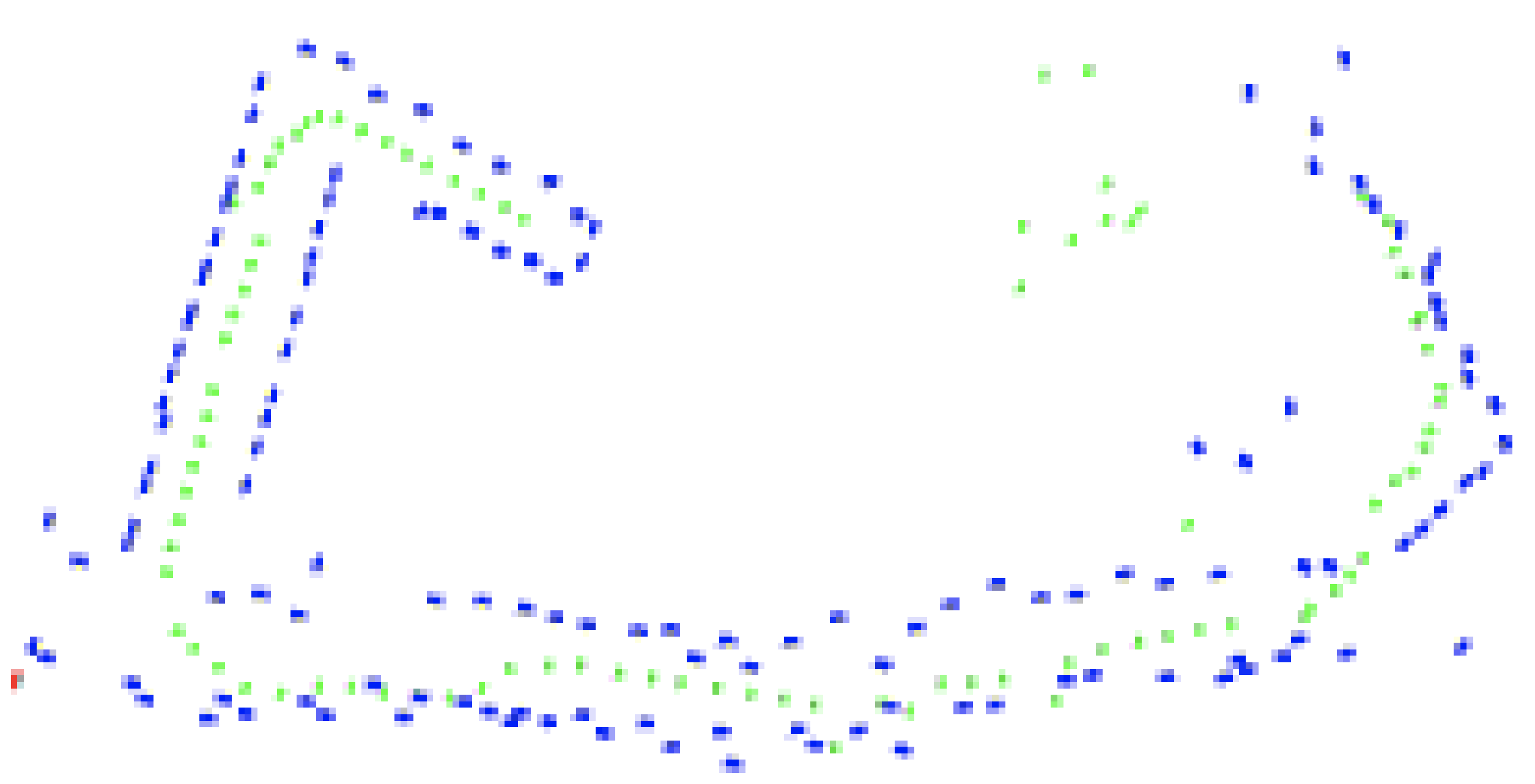}& 		
			\includegraphics[width=0.20\textwidth]{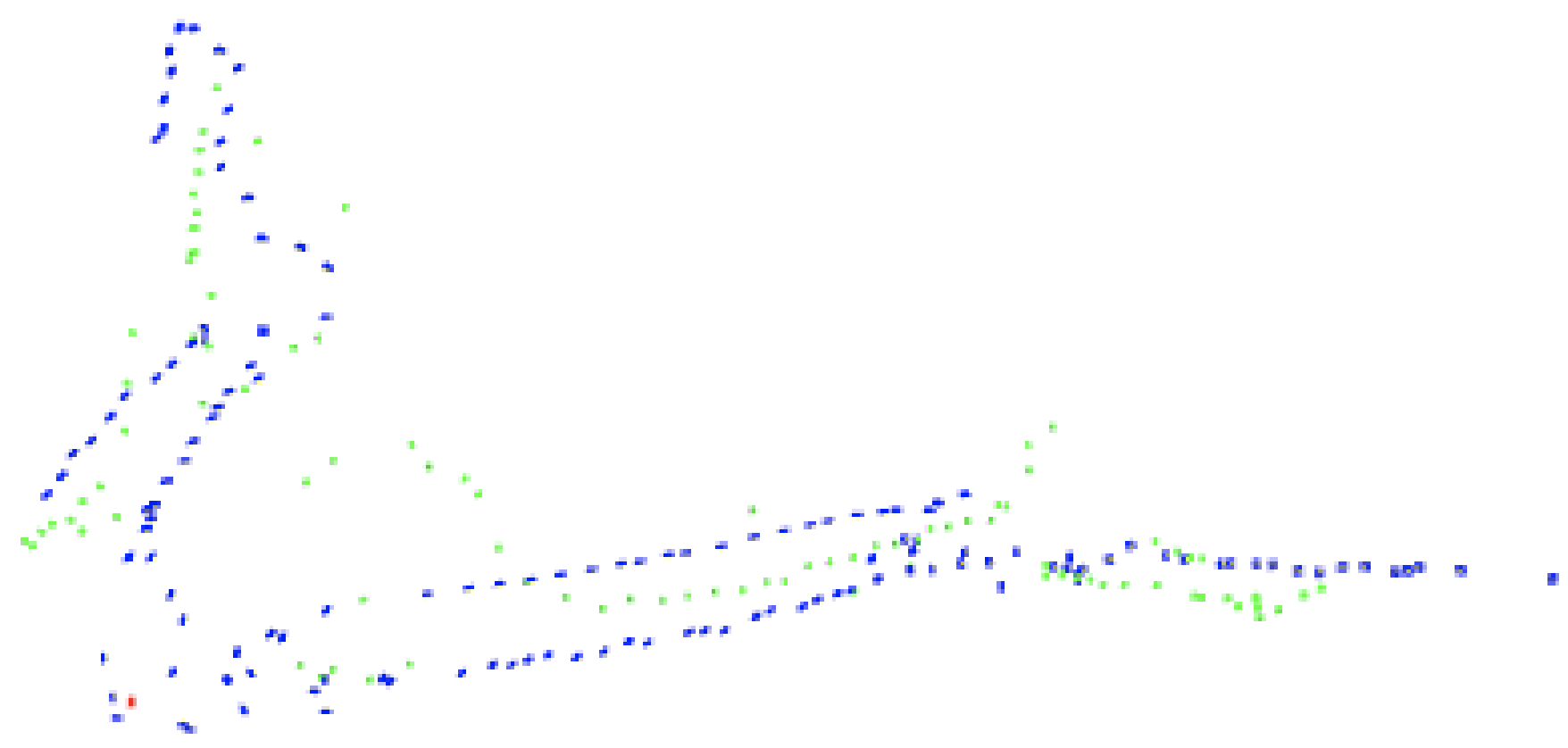}& 
			\includegraphics[width=0.20\textwidth]{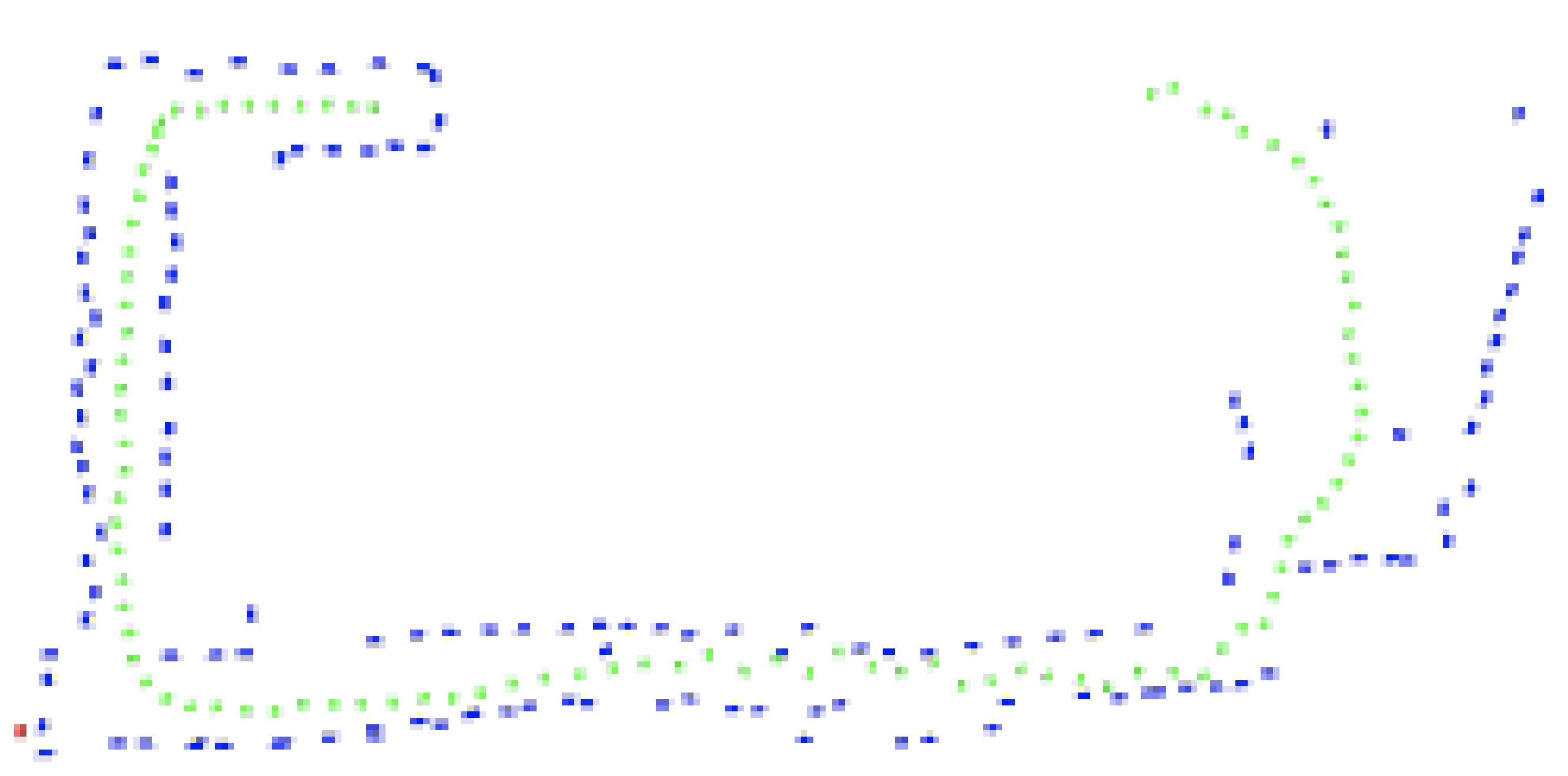} &
			\includegraphics[height=2cm,width=0.20\textwidth]{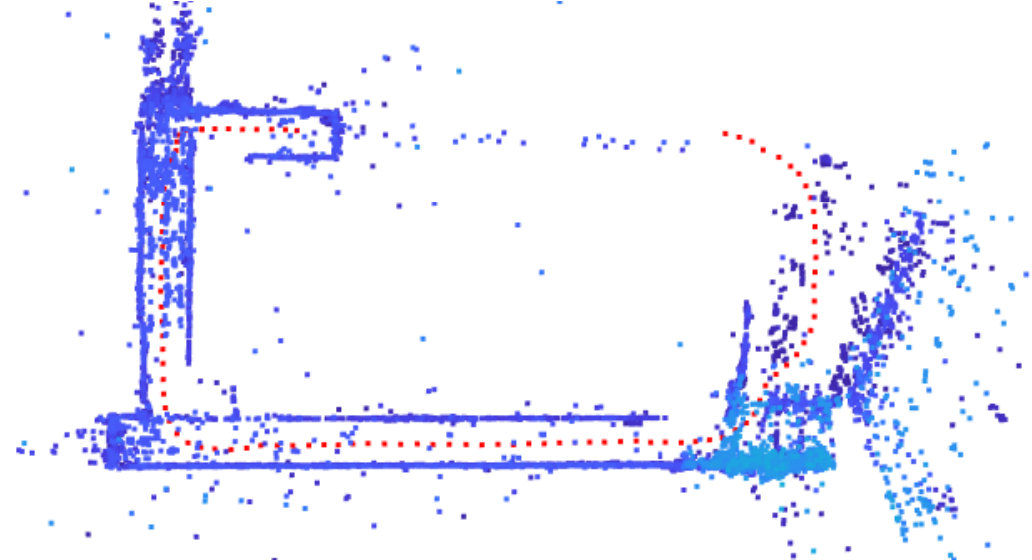}	\\
			\bottomrule 
		\end{tabular}
	}
	\caption{Qualitative result: Reconstruction results for marker-based SfM methods \textbf{M1},\textbf{M3}, \textbf{M4}, and \textbf{Ours}, as well as feature- and marker-based SfM method \textbf{FM}~\cite{degol2018improved}. Row 1: \textit{ece floor4 wall}, Row 2: \textit{ece floor5 stairs}, Row 3: \textit{cee night cw}. For the marker-based methods, red = reconstructed reference marker, blue: reconstructed markers, green: estimated camera positions.}
	\label{tab:quantitative_result}
\end{table*}

\begin{figure*}[h]
	\centering	 
	\hspace{1cm}
	\subfigure{\includegraphics[width=0.25\textwidth]{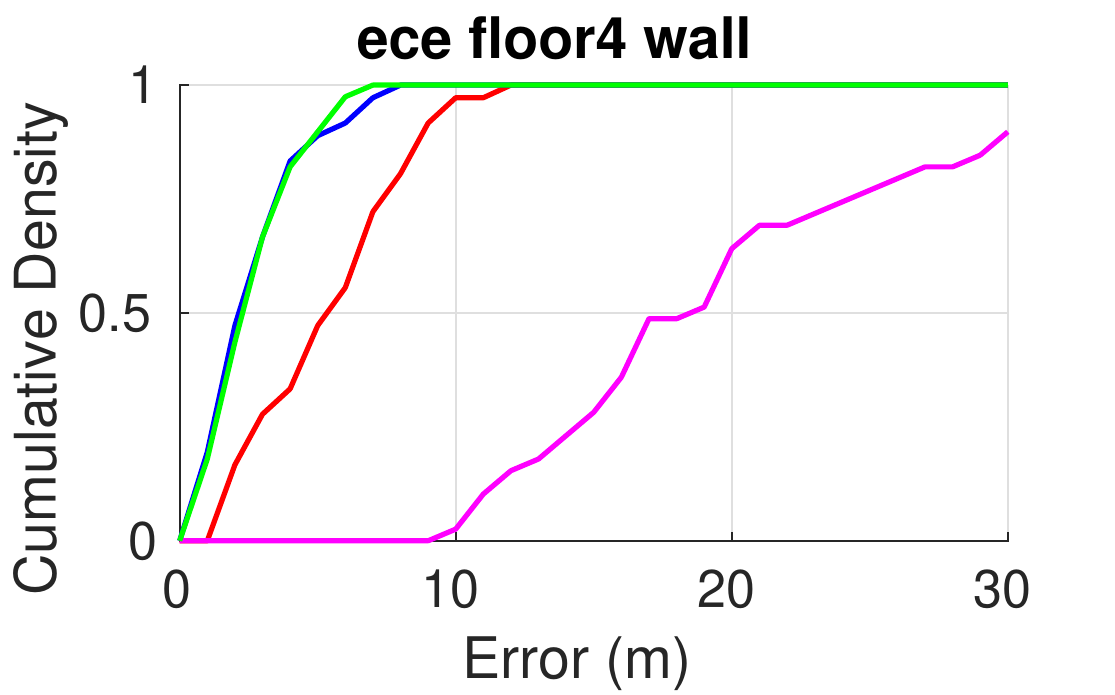}\label{subfig:ece_floor4_wall_cd}}
	\subfigure{\includegraphics[width=0.25\textwidth]{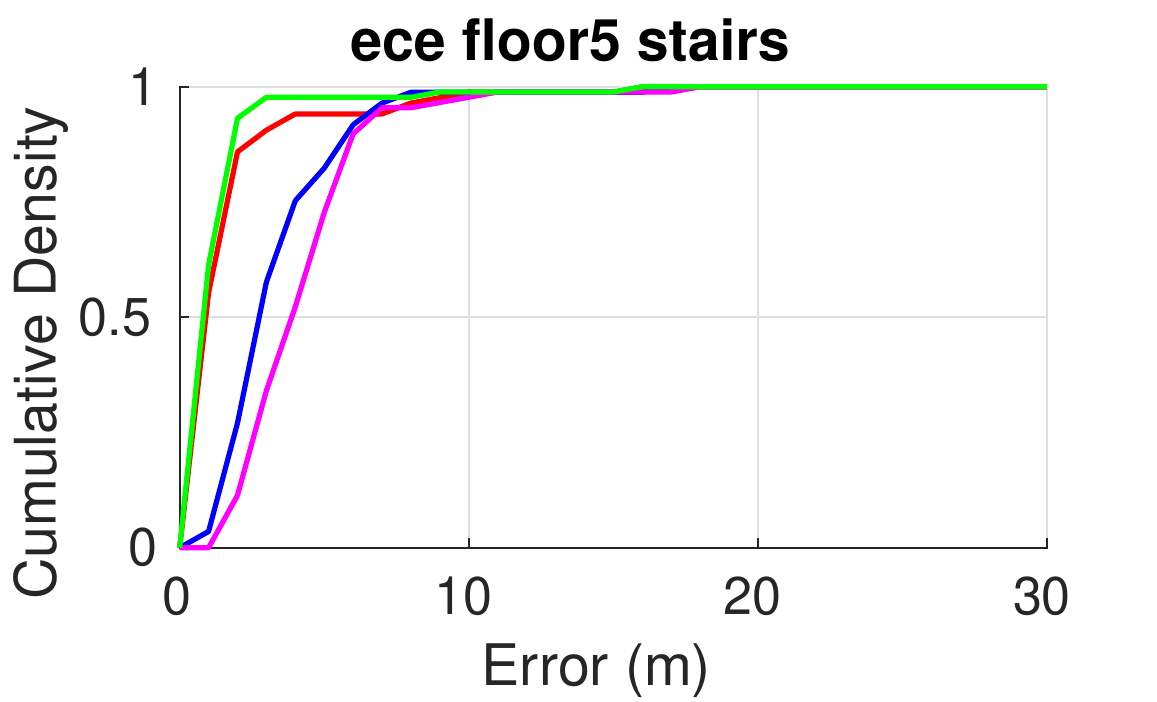}\label{subfig:ece_floor5_stairs_cd}}
	\subfigure{\includegraphics[width=0.25\textwidth]{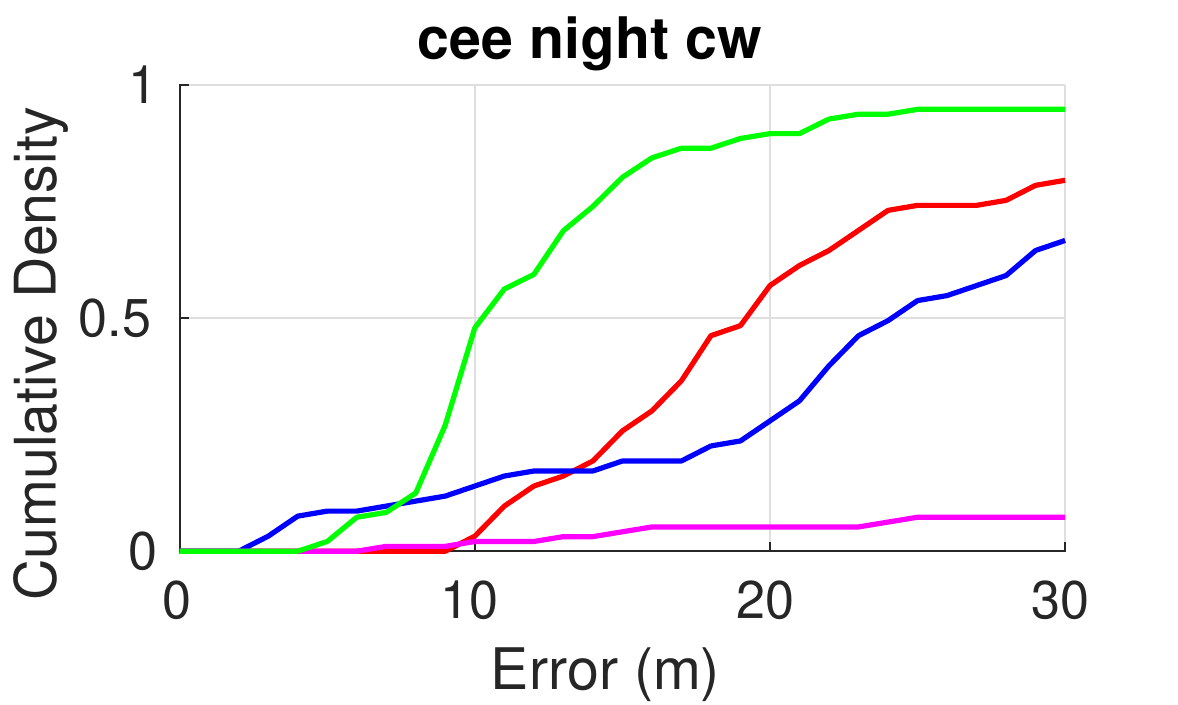}\label{subfig:cee_night_cw_cd}}
	\subfigure{\includegraphics[height=2.5cm, width=0.15\textwidth]{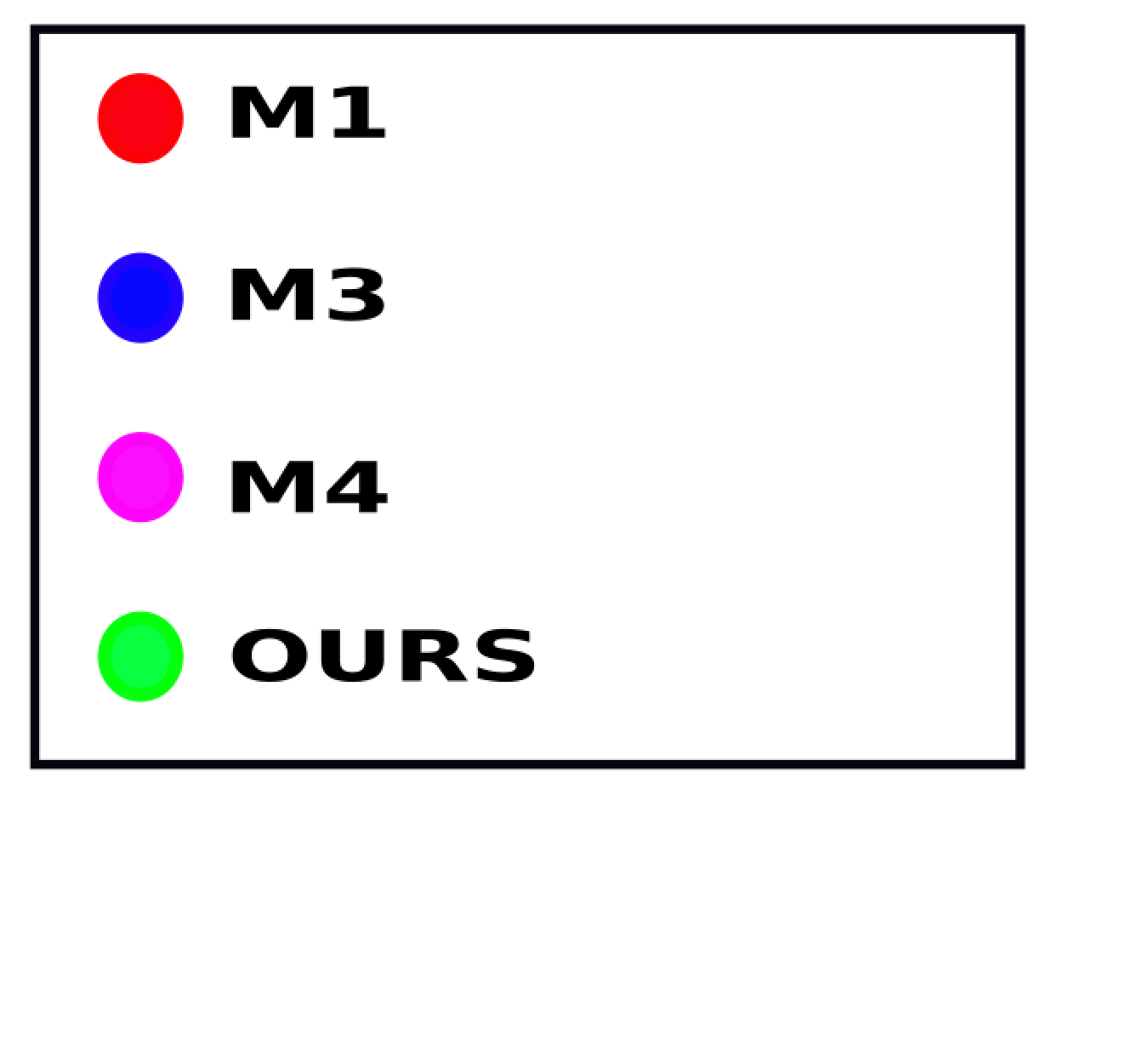}\label{subfig:cee_night_cw_cd}}
	\caption{Comparison of camera position error (relative to \textbf{FM}) of \textbf{M1, M3, M4} and \textbf{Ours}.}
	\label{fig:cumulative_density}
\end{figure*}

\begin{table}
	\centering
	\setlength{\tabcolsep}{2pt}
	\begin{tabular}{c|cccc|cccc}
		\toprule
		\textbf{Dataset}&\multicolumn{4}{c}{\textbf{Mean err. (m)}} &\multicolumn{4}{c}{\textbf{Median err. (m)}}\\
		& M1 & M3 & M4 & Ours & M1 & M3 & M4 & Ours \\ 
		\cmidrule{1-9} 
		ece floor4 wall & 5.28 & 2.72 & 20.95 & \textbf{2.56} &
		5.35 & 2.03 & 18.09 & \textbf{2.12} \\
		
		ece floor5 stairs & 1.58 & 3.18 & 4.07 & \textbf{1.14} &
		0.96 & 2.64 & 3.72 & \textbf{0.82} \\
		
		cee night cw & 30.21 & 34.79 & 75.57 & \textbf{19.06} &
		19.25 & 24.21 & 76.42 & \textbf{10.12} \\
		\bottomrule	
	\end{tabular}
	\caption{Mean and median camera position error, relative to \textbf{FM}.}\label{tab:meanmederr}
\end{table}

\subsubsection{Ground truth M2C pose selection} \label{subsec:groundtruth_label}

On the noisy corner localisations, PPE~\cite{collins2014infinitesimal} is invoked, which yields two M2C relative poses $\{\tilde \bp^{(t,a)}_i\}_{a=0,1}$ for each detected marker. To decide the ground truth selection, we compute the angular difference $\{\theta_{i}^{(t,a)}\}_{a=0,1}$ between $ \{\tilde \bR_{i}^{(t,a)}\}_{a=0,1}$ and $\bar{\bR}_i^{(t)}$ as
\begin{align}
\theta_i^{(t,a)} = \frac{180}{\pi}\,\text{acos}(1 - 0.25\, \|\,\text{I} - \tilde{\bR}_i^{(t,a)} (\bar{\bR}_i^{(t)})^T\|^2_F).
\end{align}
The ground truth selection of the PPE poses is taken as the one with the lower angular difference $\min\{\theta_{i}^{(t,a)}\}_{a=0,1}$.

\subsubsection{Results}

For the hybrid data experiment, we evaluated all the approaches on two main aspects; see supplementary video~$^{\ref{footnote:video_link}}$ for demonstration of our pose disambiguation method.

\paragraph{Precision in pose disambiguation}

For each testing sequence, precision in pose disambiguation is defined as
\begin{align}
\frac{\text{\# number of correct PPE pose selections}}{\text{\# marker detections where a decision was made}}.
\end{align}
Table~\ref{tab:synthetic_data_precision_disambiguation} shows that~\textbf{Ours} generally has higher precision than the others. The fact that~\textbf{M4} (the control method) is much poorer than \textbf{Ours} proves that enforcing the proposed clique-consistency is crucial for disambiguating the PPE poses. Amongst the per-marker disambiguation methods (\textbf{M1}--\textbf{M3}), \textbf{M1} has the lowest precision, validating observations in previous works that comparing reprojection errors alone is not foolproof. Adding a ratio test to avoid decisions on cases that are too ambiguous helps to improve precision in \textbf{M2} and \textbf{M3}. In particular, the precision of \textbf{M2} is on par with \textbf{Ours}. However, as we show next, this gain by \textbf{M2} comes at a cost.

\paragraph{Completeness and accuracy of SfM}\label{subsubsec:mapping_and_localisation_acc}

To assess the effects of marker pose disambiguation on SfM, we evaluate
\begin{itemize}[leftmargin=1em]
\item the number of markers mapped and cameras localised; and
\item the error (in deg and cm) of the marker and camera poses
\end{itemize}
estimated by marker-based SfM from the disambiguated PPE poses in Table~\ref{tab:synthetic_data_precision_disambiguation},\ref{tab:synthetic_data_sfm_accuracy} respectively.  Although \textbf{M2} is precise, it yields a much sparser map than the others; moreover, as it has pruned away many useful detections, there are insufficient data to allow accurate SfM. Using our pose disambiguation technique leads to more complete and accurate maps.

\subsection{Real world dataset experiment}

Testing was performed on sequences from~\cite{degol2018improved}. We selected 3 indoor scenes with different difficulty levels: \textit{ece floor 4 wall}, \textit{ece floor5 stairs} and \textit{cee night cw}. There are $N\geq50$ unique markers placed the scene in each sequence. To enable comparisons, we invoked~\cite{degol2018improved} (denoted as \textbf{FM}) which conducts both feature- and marker-based SfM on the sequences. Since SfM with \textbf{M2} failed in all 3 sequences due to insufficient data for optimisation, comparison is not made.

Qualitative results in Table~\ref{tab:quantitative_result} show that \textbf{Ours} is more accurate than~\textbf{M1} and \textbf{M3} in marker-based SfM - of course, \textbf{Ours} is visibly not as complete as \textbf{FM}, but the latter uses features on top of markers, which entails heavier computations. Using the estimated camera positions by \textbf{FM} as reference, we obtain the position errors (in m) computed by the marker-based SfM methods - normalised and plotted as a cumulative density in Fig.~\ref{fig:cumulative_density}. It is apparent that \textbf{Ours} is much more accurate in camera localisation, especially in the most challenging sequence~\textit{cee night cw}. Table~\ref{tab:meanmederr} lists the mean and median position error, relative to \textbf{FM}.


			
			
			


\bibliographystyle{IEEEtran}
\bibliography{IEEEabrv,egbib}

\end{document}